\documentclass[10pt,twocolumn,letterpaper]{article}

\usepackage{iccv}
\usepackage{times}
\usepackage{epsfig}
\usepackage{graphicx}
\usepackage{amsmath}
\usepackage{amssymb}
\usepackage{booktabs}
\usepackage{bbding}
\usepackage{multirow}
\usepackage[normalem]{ulem}
\useunder{\uline}{\ul}{}
\usepackage{float}
\usepackage{bm}
\makeatletter
\newcommand{\rmnum}[1]{\romannumeral #1}
\newcommand{\Rmnum}[1]{\expandafter\@slowromancap\romannumeral #1@}
\makeatother
\usepackage[table]{xcolor}
\usepackage{url}

\usepackage{utfsym}

\usepackage[pagebackref=true,breaklinks=true,letterpaper=true,colorlinks,bookmarks=false,citecolor=teal]{hyperref}

\iccvfinalcopy 

\def\iccvPaperID{1288} 

\ificcvfinal\fi

\begin{document}

\title{AIDE: A Vision-Driven Multi-View, Multi-Modal, Multi-Tasking Dataset  \\ for Assistive Driving Perception}

\author{Dingkang Yang$^{1,2 \,\,\usym{1F396}}$ $\quad$
        Shuai Huang$^{1}$\footnotemark[2] $\quad$
        Zhi Xu$^{1}$\footnotemark[2] $\quad$
        Zhenpeng Li$^{1}$\footnotemark[2] $\quad$
        Shunli Wang$^{1}$\footnotemark[2] $\quad$ \\
        Mingcheng Li$^{1}$\footnotemark[2] $\quad$ 
        Yuzheng Wang$^{1}$\footnotemark[2] $\quad$ 
        Yang Liu$^{1}$\footnotemark[2] $\quad$
        Kun Yang$^{1}$\footnotemark[2] $\quad$
        Zhaoyu Chen$^{1}$\footnotemark[2] $\quad$
        Yan Wang$^{1}$\footnotemark[2] $\quad$ \\
        Jing Liu$^{1}$\footnotemark[2] $\quad$
        Peixuan Zhang$^{5}$\footnotemark[2] $\quad$
        Peng Zhai$^{1}$\footnotemark[2] $\quad$
        Lihua Zhang$^{1,2,3,4}$\footnotemark[4] \\ 
        $^{1}$Academy for Engineering and Technology, Fudan University$\quad$
        $^{2}$Institute of Meta-Medical\\
$^{3}$Engineering Research Center of AI and Robotics, Ministry of Education, Shanghai, China\\
$^{4}$AI and Unmanned Systems Engineering Research Center of Jilin Province,
Changchun, China\\
$^{5}$Boli Technology Co., Ltd., Changchun, China\\
{\tt\small $\{$dkyang20,lihuazhang$\}$@fudan.edu.cn}
}

\maketitle
\ificcvfinal\thispagestyle{empty}\fi

\renewcommand{\thefootnote}{\fnsymbol{footnote}} 
\footnotetext[2]{These authors are second contributions. $^{\usym{1F396}}$Project lead.} 
\footnotetext[4]{Corresponding author.}

\begin{abstract}
\vspace{-0.2cm}
Driver distraction has become a significant cause of severe traffic accidents over the past decade. Despite the growing development of vision-driven driver monitoring systems, the lack of comprehensive perception datasets restricts road safety and traffic security. In this paper, we present an AssIstive Driving pErception dataset (AIDE) that considers context information both inside and outside the vehicle in naturalistic scenarios. AIDE facilitates holistic driver monitoring through three distinctive characteristics, including multi-view settings of driver and scene, multi-modal annotations of face, body, posture, and gesture, and four pragmatic task designs for driving understanding. To thoroughly explore AIDE, we provide experimental benchmarks on three kinds of baseline frameworks via extensive methods. Moreover, two fusion strategies are introduced to give new insights into learning effective multi-stream/modal representations. We also systematically investigate the importance and rationality of the key components in AIDE and benchmarks. The project link is \url{https://github.com/ydk122024/AIDE}.
\vspace{-0.5cm}
\end{abstract}


\section{Introduction}

Driving safety has been a significant concern over the past decade~\cite{downs2004traffic,klauer2006impact}, especially during the transition of automated driving technology from level 2 to 3~\cite{sae2018taxonomy}. According to the World Health Organization~\cite{world2015global}, there are approximately 1.35 million road traffic deaths worldwide each year. More alarmingly, nearly one-fifth of road accidents are caused by driver distraction that manifests in behavior~\cite{martin2019drive} or emotion~\cite{li2021spontaneous}. As a result, active monitoring of the driver's state and intention has become an indispensable component in significantly improving road safety via Driver Monitoring Systems (DMS).
Currently, vision is the most cost-effective and richest source~\cite{sivak1996information} of perception information, facilitating the rapid development of DMS~\cite{el2019driver,koesdwiady2016recent}.
Most commercial DMS rely on vehicle measures such as steering or lateral control to assess drivers~\cite{el2019driver}.
In contrast, the scientific communities~\cite{friedrichs2010camera,kasahara2022look,kopuklu2021driver,mcdonald2020classification,ortega2020dmd,zhao2018driver} focus on developing the next-generation vision-driven DMS to detect potential distractions and alert drivers to improve driving attention.
Although DMS-related datasets~\cite{statefram,eraqi2019driver,jegham2019mdad,jegham2020novel,jeong2018driver,li2021spontaneous,li2021cogemonet,martin2019drive,ortega2020dmd,saad2020end,tran2018real,zhang2020driver} offer promising prospects for enhancing driving comfort and eliminating safety hazards~\cite{mcdonald2020classification}, two serious shortcomings among them restrict the progress and application in practical driving scenarios.
\begin{table*}[t]
\centering
\caption{Comparison of public vision-driven assistive driving perception datasets. The following symbols are used in the table. 
\textbf{DBR}: driver behavior recognition; \textbf{DER}: driver emotion recognition; \textbf{TCR}: traffic context recognition; \textbf{VCR}: vehicle condition recognition; \textbf{H}: the hours of videos; \textbf{K/M}: the number of images/frames; $\ast$: the number of video clips; \textbf{N/A}: information not clarified by the authors.}    
\resizebox{\linewidth}{!}{%
\begin{tabular}{cccccccccccccc}
\toprule
Dataset    & Views & Classes & Size    & Recording Conditions & Scenarios    & Resolution        & Multimodal Annotations & DBR & DER & TCR & VCR & Usage                                                                                                     \\ \midrule
SEU~\cite{zhao2012recognition}        & 1    & 4  & 80      & Car                  & Induced      & 640 $\times$ 480         & \textbf{--}          & \CheckmarkBold& \textbf{--}  & \textbf{--}  & \textbf{--}  & Driver  postures                                                                                         \\
Tran \etal \cite{tran2018real}  & 1   & 10   & 35K     & Simulator            & Induced      & 640 $\times$ 480               & \textbf{--}                     & \CheckmarkBold& \textbf{--}  & \textbf{--}  & \textbf{--}  &  Safe driving, Distraction                                                                             \\
Zhang \etal \cite{zhang2020driver} & 2   & 9   & 60H     & Simulator            & Induced      & 640 $\times$ 360         & \CheckmarkBold                   & \CheckmarkBold& \textbf{--}  & \textbf{--}  & \textbf{--}  & Normal driving, Distraction                                                                               \\
StateFarm \cite{statefram}  & 1   & 10   & 22K     & Car                  & Induced      & 640 $\times$ 480         & \textbf{--}                     & \CheckmarkBold& \textbf{--}  & \textbf{--}  & \textbf{--}  & Normal driving, Distraction                                                                            \\
AUC-DD~\cite{eraqi2019driver}     & 1   & 10   & 14K     & Car                  & Naturalistic & 1920 $\times$ 1080 & \textbf{--}                     & \CheckmarkBold& \textbf{--}  & \textbf{--}  & \textbf{--}  &  Driver postures, Distraction \\                        
LoLi~\cite{saad2020end}       & 1    & 10  & 52K     & Car                  & Naturalistic & 640 $\times$ 480         & \CheckmarkBold                   & \CheckmarkBold& \textbf{--}  & \textbf{--}  & \textbf{--}  & Driver monitoring, Distraction                                                                            \\
Brain4Cars~\cite{jain2016brain4cars}       & 2    & 5  & 2M     & Car                  & Naturalistic & N/A         & \CheckmarkBold                   & \CheckmarkBold& \textbf{--}  & \textbf{--}  & \textbf{--}  & Driving maneuver anticipation                                                                            \\
Drive\&Act~\cite{martin2019drive} & 6   & 83   & 9.6M    & Car                  & Induced      & 1280 $\times$ 1024         & \CheckmarkBold                   & \CheckmarkBold& \textbf{--}  & \textbf{--}  & \textbf{--}  & Autonomous driving, Distraction                                                                           \\
DMD~\cite{ortega2020dmd}        & 3    & 93  & 41H   & Simulator, Car        & Induced      & 1920 $\times$ 1080         & \CheckmarkBold                   & \CheckmarkBold& \textbf{--}  & \textbf{--}  & \textbf{--}  & Distraction, Drowsiness                                                                                  \\
DAD~\cite{kopuklu2021driver}       & 2    & 24  & 2.1M    & Simulator            & Induced      & 224 $\times$ 171         & \CheckmarkBold                   & \CheckmarkBold& \textbf{--}  & \textbf{--}  & \textbf{--}  & Driver anomaly detection                                                                                 \\
DriPE~\cite{guesdon2021dripe}      & 1   & \textbf{--}   & 10K     & Car                  & Naturalistic & N/A               & \textbf{--}                     & \textbf{--}  & \textbf{--}  & \textbf{--}  & \textbf{--}  & Driver pose estimation                                                                                   \\
LBW~\cite{kasahara2022look}       & 2   & \textbf{--}   & 123K    & Car                  & Naturalistic & N/A               & \textbf{--}                     & \textbf{--}  & \textbf{--}  & \textbf{--}  & \textbf{--}  & Driver gaze estimation                 \\
MDAD~\cite{jegham2019mdad}       & 2    & 16  & 3200$^{\ast}$      & Car                  & Naturalistic      & 640 $\times$ 480         & \CheckmarkBold                   & \CheckmarkBold& \textbf{--}  & \textbf{--}  & \textbf{--}  & Driver monitoring, Distraction                                                                            \\
3MDAD~\cite{jegham2020novel}      & 2   & 16   & 574K    & Car                  & Naturalistic      & 640 $\times$ 480         & \CheckmarkBold                   & \CheckmarkBold& \textbf{--}  & \textbf{--}  & \textbf{--}  & Driver monitoring, Distraction                                                                            \\
DEFE~\cite{li2021spontaneous}       & 1   & 12   & 164$^{\ast}$  & Simulator            & Induced      & 1920 $\times$ 1080 & \textbf{--}                     & \textbf{--}  & \CheckmarkBold& \textbf{--}  & \textbf{--}  & Driver emotion understanding                                                                            \\
DEFE+~\cite{li2021cogemonet}     & 1   & 10   & 240$^{\ast}$  & Simulator            & Induced      & 640 $\times$ 480         & \CheckmarkBold                   & \textbf{--}  & \CheckmarkBold& \textbf{--}  & \textbf{--}   & Driver emotion understanding                                                                            \\
Du \etal~\cite{du2020convolution}    & 1   & 5   & 894$^{\ast}$  & Simulator            & Induced      & 1920 $\times$ 1080         & \CheckmarkBold                    & \textbf{--}  & \CheckmarkBold& \textbf{--}  & \textbf{--}  & \begin{tabular}[c]{@{}c@{}}Driver emotion understanding, \\ Biometric signal detection\end{tabular}     \\
KMU-FED~\cite{jeong2018driver}    & 1    & 6  & 1.1K    & Car                  & Naturalistic & 1600 $\times$ 1200         & \textbf{--}                     & \textbf{--}  & \CheckmarkBold& \textbf{--}  & \textbf{--}  & Driver emotion understanding                                                                            \\
MDCS~\cite{oh2022multimodal}       & 2   & 4   & 112H    & Car                  & Naturalistic & 1280 $\times$ 720               & \CheckmarkBold                   & \textbf{--}  & \CheckmarkBold& \textbf{--}  & \textbf{--}  & Driver emotion understanding                                                                            \\ \midrule \rowcolor{blue!8}
\textbf{AIDE (ours)}      & 4   & 20   & 521.64K & Car                  & Naturalistic & 1920 $\times$ 1080         & \CheckmarkBold                   & \CheckmarkBold& \CheckmarkBold& \CheckmarkBold& \CheckmarkBold& \begin{tabular}[c]{@{}c@{}c@{}}Driver monitoring, Distraction, \\ Driver emotion understanding, \\ Driving context understanding\end{tabular} \\ \bottomrule
\end{tabular}
}
\label{tab1}
\vspace{-0.3cm}
\end{table*}

We first illustrate a comprehensive comparison of mainstream vision-driven assistive driving perception datasets in Table~\ref{tab1}. Specifically, previous datasets~\cite{statefram,friedrichs2010camera,kopuklu2021driver,martin2019drive,ortega2020dmd,tran2018real,zhang2020driver,zhao2012recognition,zhao2018driver} mainly concern the in-vehicle view to observe driver-centered endogenous representations, such as 
anomaly detection~\cite{kopuklu2021driver}, drowsiness prediction~\cite{friedrichs2010camera,zhao2018driver}, and distraction recognition~\cite{statefram,tran2018real,zhang2020driver}.
However, the equally important exogenous scene factors that cause driver distraction are usually ignored. The driver's state inside the vehicle is frequently closely correlated with the traffic scene outside the vehicle~\cite{roth2016driver,zabihi2017detection}.
For instance, the reason for an angry driver to look around is most likely due to a traffic jam or malicious overtaking~\cite{kotseruba2022attention}.
Meanwhile, most smoking or talking behaviors occur in smooth traffic conditions. A holistic understanding of driver performance, vehicle condition, and scene context is imperative and promising for achieving more effective assistive driving perception.

Another shortcoming is that most existing datasets~\cite{eraqi2019driver,jegham2020novel,kopuklu2021driver,martin2019drive,ortega2020dmd,saad2020end} focus on identifying driver behavior characteristics while neglecting to evaluate their emotional states.
Driver emotion plays an essential role in complex driving dynamics as it inevitably affects driver behavior and road safety~\cite{li2020influence}.
Many researchers~\cite{cai2011modeling,russell1980circumplex} have indicated that drivers with peaceful emotions tend to maintain the best driving performance (\ie, \emph{normal driving}). Conversely, negative emotional states (\eg, \emph{weariness}) are more likely to induce distractions and secondary behaviors (\eg, \emph{dozing off})~\cite{jeon2014effects}.
Despite initial progress in driving emotion understanding works~\cite{du2020convolution,jeong2018driver,li2021spontaneous,li2021cogemonet}, these inadequate efforts only consider facial expressions and ignore the valuable clues provided by the body posture and scene context~\cite{yang2023context,yang2022disentangled,yang2022contextual,yang2022emotion,yang2022learning,yang2023target}. Most importantly, there are no comprehensive datasets that simultaneously consider the complementary perception information among driver behavior, emotion, and traffic context, which potentially limits the improvement of the next-generation DMS.

Motivated by the above observations, we propose an AssIstive Driving pErception dataset (AIDE) to facilitate further research on the vision-driven DMS. 
AIDE captures rich information inside and outside the vehicle from several drivers in realistic driving conditions.
As shown in Figure~\ref{overview}, we assign AIDE three significant characteristics.
\textbf{(i) Multi-view}: four distinct camera views provide an expansive perception perspective, including three out-of-vehicle views to observe the traffic scene context and an in-vehicle view to record the driver's state.
\textbf{(ii) Multi-modal}: diverse data annotations from the driver support comprehensive perception features, including face, body, posture, and gesture information. \textbf{(iii) Multi-task}: four pragmatic driving understanding tasks guarantee holistic assistive perception, including driver-centered behavior and emotion recognition, traffic context, and vehicle condition recognition.

To systematically evaluate the challenges brought by AIDE, we implement three types of baseline frameworks using representative and impressive methods, which involve classical, resource-efficient, and state-of-the-art (SOTA) backbone models. Diverse benchmarking frameworks provide sufficient insights to specify suitable network architectures for real-world driving perception.
For multi-stream/modal inputs, we design adaptive and cross-attention fusion modules to learn effectively shared representations. Additionally, numerous ablation studies are performed to thoroughly demonstrate the effectiveness of key components and the importance of AIDE.

\section{Related Work}
\subsection{Vision-driven Driver Monitoring Datasets}
Vision-driven driver monitoring aims to observe features from driver-related areas to identify potential distractions through various assistive driving perception tasks. According to~\cite{ortega2020dmd}, existing datasets can be categorized as follows.

\noindent\textbf{Hands-focused Datasets}. Hand poses are an important basis for evaluating human-vehicle interaction in driving scenarios, as hands off the steering wheel are closely related to many secondary behaviors (\eg, \textit{smoking}).
These datasets generally provide annotated bounding boxes for the hands, including CVRR-HANDS 3D~\cite{ohn2013driver}, VIVA-Hands~\cite{das2015performance}, and DriverMHG~\cite{kopuklu2020drivermhg}.
Furthermore, Ohn-bar~\etal~\cite{ohn2013vehicle} collect a dataset of hand activity and posture images under different illumination settings to identify the driver's state.

\noindent\textbf{Face-focused Datasets}. The face and head provide valuable clues to observe the driver's degree of drowsiness and distraction~\cite{sikander2018driver}. There are several efforts that offer eye-tracking annotations to estimate the direction of the driver's gaze and position of attention, such as DrivFace~\cite{diaz2016reduced}, DADA~\cite{fang2019dada}, and LBW~\cite{kasahara2022look}. Some multimodal datasets~\cite{ortega2020dmd,zhang2020driver} utilize facial information as a complementary perceptual stream. Moreover, DriveAHead~\cite{schwarz2017driveahead} and DD-Pose~\cite{roth2019dd} focus on fine-grained head analysis through pose annotations of yaw, pitch, and roll angles.

\noindent\textbf{Body-focused Datasets}. Observing the driver's body actions via the in-vehicle view has become a widely adopted monitoring paradigm.
These perceptual patterns from the driver's body contain diverse resources such as keypoints~\cite{guesdon2021dripe}, RGB~\cite{tran2018real}, infrared~\cite{saad2020end}, and depth information~\cite{kopuklu2021driver}. This technical route is first led by the StateFarm~\cite{statefram} competition dataset, which contains behavioral categories of safe driving and distractions. Since then, numerous databases have been proposed to progressively enrich body-based monitoring methods. These include AUC-DD~\cite{eraqi2019driver}, Loli~\cite{saad2020end}, MDAD~\cite{jegham2019mdad}, 3MDAD~\cite{jegham2020novel}, and DriPE~\cite{guesdon2021dripe}.
More recently, some compounding efforts have considered extracting additional information, such as vehicle interiors~\cite{martin2019drive}, objects~\cite{ortega2020dmd}, and optical flow~\cite{zhang2020driver}.

We show a specification comparison with the relevant assistive driving perception datasets for the proposed AIDE.
As shown in Table~\ref{tab1}, previous datasets either deal with specific perception tasks or only focus on driver-related characteristics.
In contrast, AIDE considers the rich context clues inside and outside the vehicle and supports the collaborative perception of driver behavior, emotion, traffic context, and vehicle condition. AIDE is more multi-purpose, diverse, and holistic for assistive driving perception.

\subsection{Driving-aware Network Architectures}
DMS-oriented models usually adopt network structures that are convenient to deploy on-road vehicles.
With advances in deep learning techniques~\cite{chen2023query,chen2023content,chen2022shape,chen2022towards,du2021learning,jiang2023efficient,kuang2023towards,liu2022efficient,liu2022collaborative,liu2022learning,liu2023generalized,sun2023human,wang2022boosting,wang2022spacenet,wang2021tsa,wang2021survey,wang2023model,wang2023adversarial,xu2023model,xu2022bridging,xu2023v2v4real,yang2023novel,zhu2023direct}, most approaches that accompany datasets prioritize implementing classical models. These widely accepted network architectures include AlexNet~\cite{krizhevsky2017imagenet}, GoogleNet~\cite{szegedy2015going}, VGG~\cite{simonyan2014very}, and ResNet~\cite{he2016deep} families. Meanwhile, lightweight models with resource-efficient advantages are also favored enough, such as MobileNet~\cite{howard2017mobilenets,sandler2018mobilenetv2} and ShuffleNet~\cite{ma2018shufflenet,zhang2018shufflenet}.
3D-CNN models such as C3D~\cite{tran2015learning}, I3D~\cite{carreira2017quo}, and 3D-ResNet~\cite{hara2018can} have been implemented to capture spatio-temporal features in video-based data. 
Several tailored structures have also been presented to suit specific data patterns~\cite{ma2022real,zhang2020driver}.
We fully exploit the classical, lightweight, and SOTA baselines to implement extensive experiments across various learning paradigms. The diverse combinations of models for different input streams provide valuable insights into the appropriate structure selection.

\subsection{Driving-aware Fusion Strategies}
Various fusion strategies are proposed to meet multi-stream/modal input requirements in driving perception. The mainstream fusion patterns are divided into data-level, feature-level, and decision-level. For example, Ortega~\etal~\cite{ortega2020dmd} perform a data-level fusion of infrared and depth frames based on pixel-wise correlation to achieve better perception performance than unimodality. The common feature-level fusion is based on feature summation or concatenation~\cite{wu2021learning}. Moreover, Kopukl~\etal~\cite{kopuklu2021driver} train a separate model for each view from the driver and then achieve decision-level fusion based on similarity scores.
Here, we introduce two fusion modules at the feature level to learn effective representations among multiple feature streams.
 
 \begin{figure}[t]
  \centering
  \includegraphics[width=\linewidth]{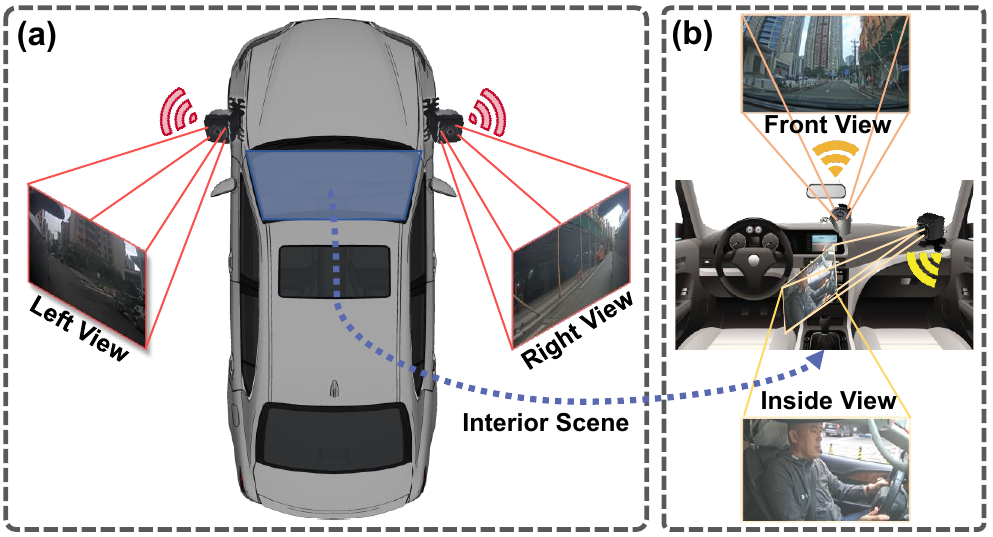}
  \caption{Camera setup for AIDE in the real vehicle scenario. The setup involves (a) exterior and (b) interior camera layouts.
  }
  \label{layout}
  \vspace{-0.4cm}
\end{figure}

\section{The AIDE Dataset}
\subsection{Data Collection Specification}
To tackle the lack of perceptually comprehensive driver monitoring benchmarks, we collect the AIDE dataset under the consecutive manual driving mode, which is essential for the transition of automated vehicles from level 2 to 3~\cite{sae2018taxonomy}. 

\noindent\textbf{Camera Setup}. The driving environment and camera layout are shown in Figure~\ref{layout}. Specifically, the experimental vehicle is used on real roads to capture rich information about the interior and exterior of the vehicle. The primary data source is four Axis cameras with 1920$\times$1080 resolution. The frame rate is 15 frames per second, and the dynamic range is 120 dB. 
Concretely, a camera is mounted in front of the vehicle's each side mirror to produce a left and right view capturing the traffic context. Meanwhile, the front view camera is mounted in the dashboard's centre to observe the front scene. For the inside view, we record the driver's natural reactions from the side in a non-intrusive way, with a clear perspective of the face, body, and hands interacting with the steering wheel.
The four connected cameras are synchronized via the Precision Timing Protocol.

\noindent\textbf{Collection Programme}. Naturalistic driving data is collected from several drivers with different driving styles and habits to ensure the authenticity of AIDE. Unlike previous efforts~\cite{jegham2019mdad,jegham2020novel,martin2019drive,ortega2020dmd} to force subjects to perform specific tasks/training to induce distraction, our data is derived from the most realistic driving performance of drivers who are not informed in advance.
The guideline aims to bridge the driving reaction gap between the experimental domain and the realistic monitoring domain. In this case, each participant's driving operation is conducted at different times on different days to contain diverse driving scenarios. From Figure~\ref{overview}, these scenario factors include distinct light intensities, weather conditions, and traffic contexts, increasing the challenge and diversity of AIDE.

 \begin{figure}[t]
  \centering 
  \includegraphics[width=\linewidth]{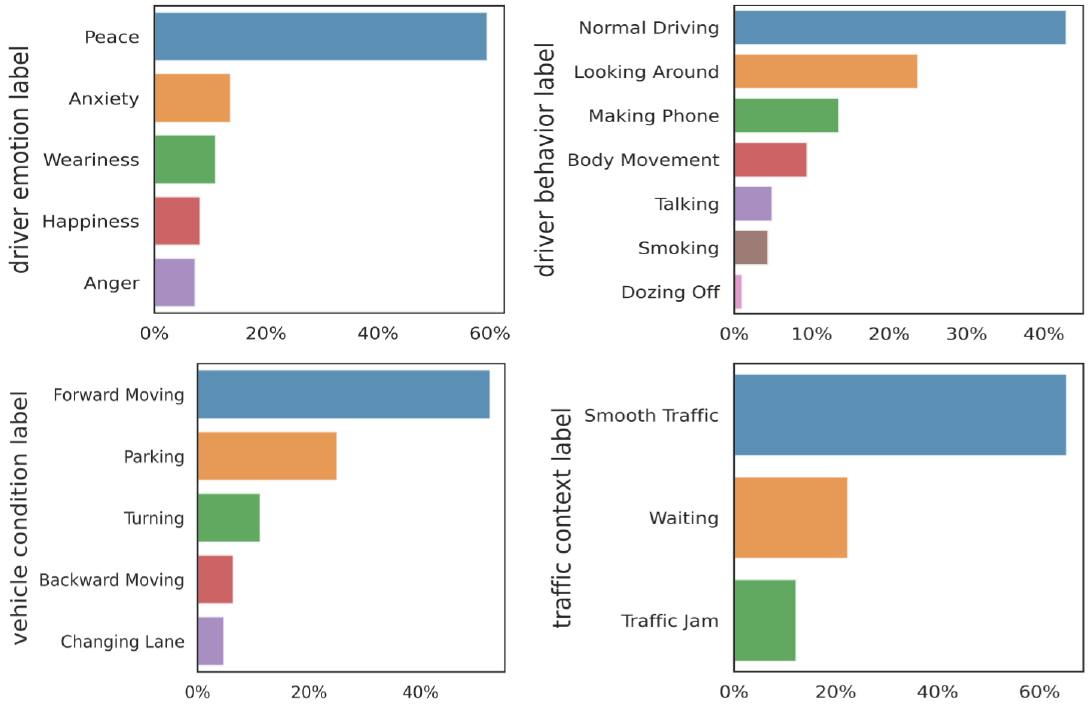}
  \caption{The percentage of samples in each category for the four driving perception tasks.
  }
  \label{count}
  \vspace{-0.3cm}
\end{figure}

\subsection{Data Stream Recording and Annotation}
\noindent\textbf{Recorded Data Streams}. 
Our AIDE has various information types to provide rich data resources for different downstream tasks, including face, body, and traffic context (\ie, out-of-vehicle views) video data, and keypoint information. As the duration of the different driving reactions varies, the raw video data from the four views are first synchronously processed into 3-second short video clips using the Moviepy Library. The processing facilitates the AIDE-based monitoring system to satisfy real-time responses within a fixed span.
For the inside view of Figure~\ref{overview}(b), the face detector MTCNN~\cite{zhang2016joint} is utilized to capture the driver's facial bounding box.
Meanwhile, the pose estimator AlphaPose~\cite{alphapose} is employed to obtain driver-centred information, including the body bounding box, 2D skeleton posture (26 keypoints), and gesture (42 keypoints). We eliminate clips with missing results based on the above detection to ensure data integrity.
An additional operation in the retained clips is applied to fill missing joints using interpolation of adjacent frames.

\noindent\textbf{Task Determination}. 
Four pragmatic assistive driving tasks are proposed to facilitate holistic perception. Endogenous Driver Behavior and Emotion Recognition (DBR, DER) are adopted because these two tasks intuitively reflect distraction/inattention~\cite{kopuklu2021driver, li2021spontaneous}. Exogenously, Traffic Context Recognition (TCR) is considered since the scene context provides valuable evidence for understanding driver intention~\cite{roth2016driver}. Also, we establish Vehicle Condition Recognition (VCR) as the driver's state usually accompanies a transition in vehicle control~\cite{kotseruba2022attention}. These complementary tasks all benefit from the rich data resources from AIDE.

\noindent\textbf{Label Assignment}. 
The dataset annotation involves 12 professional data engineers with bespoke training. The annotation is performed
blindly and independently, and we utilize the majority voting rule to determine
the final labels.
To adequately represent real driving situations, the behavior categories consist of one safe \emph{normal driving} and six secondary activities that frequently cause traffic accidents. For emotions, five categories that occur frequently and tend to induce distractions in drivers are considered.
Meanwhile, six research experts in human-vehicle interaction are asked to rate three traffic context categories and five vehicle condition categories.
Figure~\ref{overview}(c) displays each category from the different tasks and provides a corresponding illustration.

\noindent\textbf{Data Statistic}.  
Eventually, we obtained 2898 data samples with 521.64K frames.
Each sample consists of 3-second video clips from four views, where the duration shares a specific label from each perception task.
The inside clips contain the estimated bounding boxes and keypoints on each frame.
AIDE is randomly divided into training (65\%), validation (15\%), and testing (20\%) sets without considering held-out subjects due to the naturalistic nature of data imbalance. A stratified sampling is applied to ensure that each set contains samples from all categories for different tasks. Figure~\ref{count} shows the percentage of samples in each category for each task.

\noindent\textbf{Ethics Statement}. All our materials adhere to ethical standards for responsible research practice. Each participant signed a GDPR\footnote{\url{https://gdpr-info.eu/}} informed consent which allows the dataset to be publicly available for research purposes.

\section{Assistive Driving Perception Framework}
\subsection{Model Zoo}
\begin{figure}[t]
  \centering 
  \includegraphics[width=\linewidth]{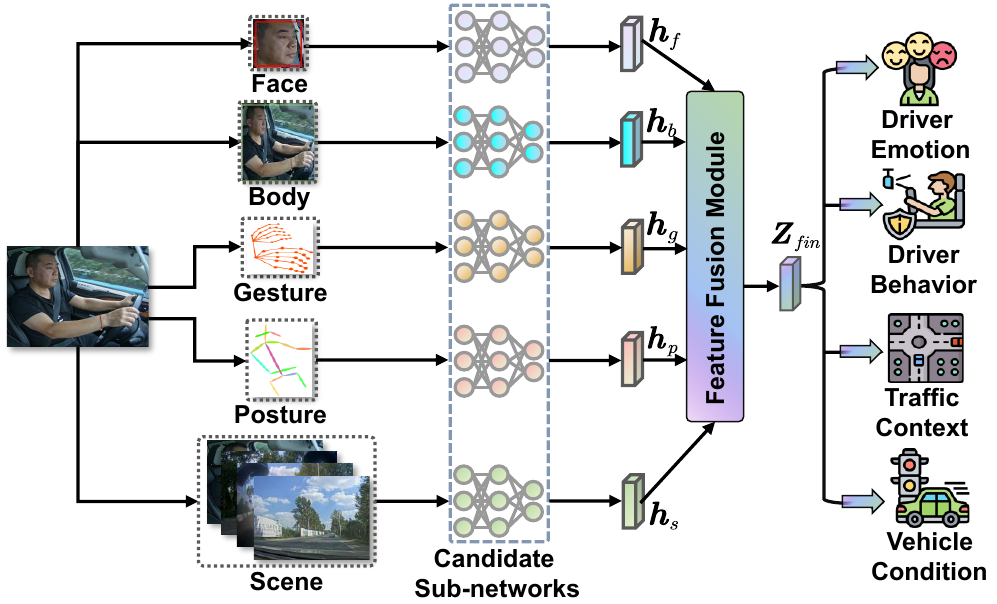}
  \caption{Our assistive driving perception framework pipeline.}
  \label{fusion}
  \vspace{-0.3cm}
\end{figure}

To thoroughly explore AIDE, we introduce three types of baseline frameworks to cover most driving perception modeling paradigms via extensive methods.
As Figure~\ref{fusion} shows, our frameworks accommodate all available streams, including video information of the face, body, and scene, as well as keypoints of gesture and posture.

\noindent\textbf{2D Pattern}. 
Classical 2D ConvNets such as ResNet~\cite{he2016deep} and VGG~\cite{simonyan2014very} have significantly succeeded in image-based recognition. Here, we reuse them with minimal change. For processing a clip, the hidden features of sampled frames are extracted simultaneously and then aggregated by a 1D convolutional layer. For the skeleton keypoints, we design Multi-Layer Perceptrons (MLPs) with GeLU~\cite{hendrycks2016gaussian} activation to perform feature extraction. Meanwhile, a Spatial Embedding (SE) is also added to provide location information.

\noindent\textbf{2D + Timing Pattern}. This pattern aims to introduce an additional sequence model after 2D ConvNets to learn temporal representations. As a result, a Transformer Encoder (TransE)~\cite{vaswani2017attention} is employed to refine the hidden features among sampled frames and then aggregated by a temporal convolutional layer. Furthermore, we augment a Temporal Embedding (TE) for the MLPs to maintain the temporal dynamics of the gesture and posture modalities.

\noindent\textbf{3D Pattern}. 
The 3D network structures directly model hierarchical representations by capturing spatio-temporal information. We consider various impressive models, including 3D-ResNet~\cite{hara2018can}, C3D~\cite{tran2015learning}, I3D~\cite{carreira2017quo}, SlowFast~\cite{feichtenhofer2019slowfast}, and TimeSFormer~\cite{bertasius2021space}.
Furthermore, the 3D versions of lightweight networks such as MobileNet-V1/V2\,\cite{howard2017mobilenets,sandler2018mobilenetv2} and ShuffleNet-V1/V2\,\cite{zhang2018shufflenet,ma2018shufflenet}, which are resource-efficient for DMS, are also considered.
In this case, we introduce the remarkable ST-GCN~\cite{yan2018spatial} to process the skeleton sequences via multi-level spatio-temporal graphs.
\subsection{Feature Fusion and Learning Strategies}
How to effectively fuse the multi-stream/modal features extracted by the above candidate networks is crucial for diverse perception tasks.
To this end, we propose two sophisticated feature-level fusion modules to learn valuable shared representations among multiple features.

\begin{table*}[t]
\setlength{\tabcolsep}{4pt}
\centering
\caption{Comparison results of baseline models in three distinct patterns on the AIDE for four tasks. In each pattern, the best results are marked in \textbf{bold}, and the second-best results are marked {\ul underlined}. The following abbreviations are used. \textbf{Res}: ResNet~\cite{he2016deep}; \textbf{MLP}: multi-layer perception; \textbf{SE}: spatial embedding; \textbf{TE}: temporal embedding; \textbf{TransE}: transformer encoder~\cite{vaswani2017attention}; \textbf{PP}: pre-training on the Places365~\cite{zhou2017places} dataset; \textbf{CG}: coarse-grained.
}
\resizebox{\linewidth}{!}{%
\begin{tabular}{c|cccccl|cccc|cccc|cc|cc|c}
\toprule
\multirow{2}{*}{Pattern}                                              & \multicolumn{6}{c|}{Backbone}                                                                   & \multicolumn{4}{c|}{DER}                   & \multicolumn{4}{c|}{DBR}                  & \multicolumn{2}{c|}{TCR} & \multicolumn{2}{c|}{VCR} & \multirow{2}{*}{ID} \\ \cline{2-19}
& Face             & Body             & Gesture & Posture & \multicolumn{2}{c|}{Scene}            & CG-Acc         & CG-F1           & Acc          & F1          & CG-Acc         & CG-F1          & Acc         & F1          & Acc                      & F1                     & Acc                     & F1                     &                     \\ \midrule
\multirow{5}{*}{2D}                                                   & Res18~\cite{he2016deep}         & Res34            & MLP+SE  & MLP+SE  & \multicolumn{2}{c|}{PP-Res18~\cite{zhou2017places}}         & 71.08          & 67.54          & 69.05          & 63.06          & {\ul 74.84}    & {\ul 74.92}    & {\ul 63.87}    & {\ul 59.52}    & \textbf{88.01}          & \textbf{86.63}         & \textbf{78.16}           & \textbf{77.27}          & (1)                   \\
 & Res18            & Res34            & MLP+SE  & MLP+SE  & \multicolumn{2}{c|}{Res34}            & \textbf{73.23} & \textbf{70.47} & \textbf{71.26} & \textbf{68.71} & \textbf{75.37} & \textbf{75.58} & \textbf{65.35} & \textbf{63.29} & {\ul 83.74}             & {\ul 81.28}            & 77.12                    & 75.23                   & (2)                   \\
& Res34            & Res50            & MLP+SE  & MLP+SE  & \multicolumn{2}{c|}{Res50}            & 72.62          & 68.75          & 69.68          & 64.83          & 73.01          & 72.75          & 59.77          & 54.64          & 80.13                   & 74.47                  & 71.26                    & 69.53                   & (3)                   \\
 & VGG13~\cite{simonyan2014very}            & VGG16            & MLP+SE  & MLP+SE  & \multicolumn{2}{c|}{VGG16}            & {\ul 73.15}    & {\ul 70.25}    & {\ul 70.72}    & {\ul 67.11}    & 74.71          & 74.61          & 63.65          & 58.12          & 82.77                   & 80.42                  & {\ul 77.94}              & {\ul 76.29}             & (4)                   \\
 & VGG16            & VGG19            & MLP+SE  & MLP+SE  & \multicolumn{2}{c|}{VGG19}            & 71.23          & 67.79          & 69.31          & 64.67          & 72.66          & 72.73          & 62.34          & 57.33          & 83.58                   & 80.67                  & 75.13                    & 73.96                   & (5)                   \\ \midrule
\multirow{5}{*}{\begin{tabular}[c]{@{}c@{}}2D +\\ Timing\end{tabular}} & Res18+TransE     & Res34+TransE     & MLP+TE  & MLP+TE  & \multicolumn{2}{c|}{PP-Res18+TransE}  & 73.28          & 71.29          & 70.83          & 67.14          & \textbf{76.44} & \textbf{76.86} & \textbf{67.32} & \textbf{64.45} & \textbf{90.54}          & \textbf{89.66}         & \textbf{79.97}           & \textbf{77.94}          & (6)                   \\
& Res18+TransE     & Res34+TransE     & MLP+TE  & MLP+TE  & \multicolumn{2}{c|}{Res34+TransE}     & \textbf{75.37} & \textbf{74.68} & \textbf{72.65} & \textbf{70.96} & 76.35          & 76.77          & 67.08          & 64.11          & {\ul 86.63}             & {\ul 84.87}            & 78.46                    & 76.51                   & (7)                   \\
  & Res34+TransE     & Res50+TransE     & MLP+TE  & MLP+TE  & \multicolumn{2}{c|}{Res50+TransE}     & 72.89          & 69.06          & 70.24          & 65.65          & 74.28          & 74.32          & 63.54          & 59.91          & 82.57                   & 77.29                  & 73.69                    & 72.26                   & (8)                   \\
 & VGG13+TransE     & VGG16+TransE     & MLP+TE  & MLP+TE  & \multicolumn{2}{c|}{VGG16+TransE}     & {\ul 74.55}    & {\ul 73.45}    & {\ul 71.12}    & {\ul 69.58}    & {\ul 76.37}    & {\ul 76.81}    & {\ul 67.15}    & {\ul 64.27}    & 85.13                   & 83.34                  & {\ul 78.58}              & {\ul 76.77}             & (9)                   \\
& VGG16+TransE     & VGG19+TransE     & MLP+TE  & MLP+TE  & \multicolumn{2}{c|}{VGG19+TransE}     & 72.57          & 68.39          & 69.46          & 64.75          & 73.71          & 73.48          & 65.48          & 61.71          & 85.74                   & 83.95                  & 77.91                    & 76.05                   & (10)                  \\ \midrule
\multirow{10}{*}{3D}                                                  & MobileNet-V1~\cite{howard2017mobilenets}  & MobileNet-V1  & ST-GCN   & ST-GCN   & \multicolumn{2}{c|}{MobileNet-V1}  & 74.71          & 73.47          & 72.23          & 69.61          & 75.04          & 75.26          & 64.20           & 61.48          & 88.34                   & 86.95                  & 77.83                    & 75.69                   & (11)                  \\
& MobileNet-V2~\cite{sandler2018mobilenetv2}  & MobileNet-V2  & ST-GCN   & ST-GCN   & \multicolumn{2}{c|}{MobileNet-V2}  & 70.27          & 66.54          & 68.47          & 62.58          & 70.28          & 69.98          & 61.74          & 54.74          & 86.54                   & 82.38                  & 78.66                    & 76.78                   & (12)                  \\
  & ShuffleNet-V1~\cite{zhang2018shufflenet} & ShuffleNet-V1 & ST-GCN   & ST-GCN   & \multicolumn{2}{c|}{ShuffleNet-V1} & {\ul 75.21}    & {\ul 74.44}    & {\ul 72.41}    & {\ul 70.82}    & {\ul 76.19}    & {\ul 76.36}    & \textbf{68.97} & \textbf{67.13} & {\ul 90.64}             & {\ul 89.98}            & {\ul 80.79}              & {\ul 79.66}             & (13)                  \\
 & ShuffleNet-V2~\cite{ma2018shufflenet} & ShuffleNet-V2 & ST-GCN   & ST-GCN   & \multicolumn{2}{c|}{ShuffleNet-V2} & 74.38          & 73.42          & 70.94          & 69.53          & 73.56          & 73.78          & 64.04          & 61.75          & 89.33                   & 87.54                  & 78.98                    & 77.52                   & (14)                  \\
& 3D-Res18~\cite{hara2018can}         & 3D-Res34         & ST-GCN   & ST-GCN   & \multicolumn{2}{c|}{3D-Res34}         & 73.07          & 70.23          & 70.11          & 65.15          & \textbf{78.16} & \textbf{78.35} & {\ul 66.52}    & {\ul 64.57}    & 88.51                   & 87.26                  & \textbf{81.12}           & \textbf{79.71}          & (15)                  \\
  & 3D-Res34         & 3D-Res50         & ST-GCN   & ST-GCN   & \multicolumn{2}{c|}{3D-Res50}         & 70.61          & 67.10           & 69.13          & 62.95          & 71.26          & 71.01          & 63.05          & 57.97          & 87.82                   & 84.86                  & 79.31                    & 76.87                   & (16)                  \\
& C3D~\cite{tran2015learning}              & C3D              & ST-GCN   & ST-GCN   & \multicolumn{2}{c|}{C3D}              & 66.35          & 62.04          & 63.05          & 57.06          & 73.57          & 73.64          & 63.95          & 60.36          & 85.41                   & 80.44                  & 77.01                    & 74.84                   & (17)                  \\
 & I3D~\cite{carreira2017quo}              & I3D              & ST-GCN   & ST-GCN   & \multicolumn{2}{c|}{I3D}              & 71.43          & 68.05          & 70.94          & 65.99          & 74.38          & 74.36          & 66.17          & 61.35          & 87.68                   & 84.78                  & 79.81                    & 78.66                   & (18)                  \\
  & SlowFast~\cite{feichtenhofer2019slowfast}         & SlowFast         & ST-GCN   & ST-GCN   & \multicolumn{2}{c|}{SlowFast}         & 75.17          & 74.24          & 72.38          & 70.77          & 75.53          & 75.73          & 61.58          & 59.41          & 86.86                   & 84.66                  & 78.33                    & 76.66                   & (19)                  \\
   & TimeSFormer~\cite{bertasius2021space}      & TimeSFormer      & ST-GCN   & ST-GCN   & \multicolumn{2}{c|}{TimeSFormer}      & \textbf{76.52} & \textbf{74.92} & \textbf{74.87} & \textbf{72.56} & 73.73          & 73.91          & 65.18          & 63.24          & \textbf{92.12}          & \textbf{91.81}         & 78.81                    & 76.91                   & (20)                  \\ \bottomrule
\end{tabular}
}
\label{tab2}
\vspace{-0.4cm}
\end{table*}
\noindent\textbf{Adaptive Fusion Module}.
Modality heterogeneity leads to distinct features contributing differently to the final prediction. The adaptive fusion module aims to assign dynamic weights to target features $\bm{F}_{ta} \in \{\bm{h}_f, \bm{h}_b, \bm{h}_g, \bm{h}_p, \bm{h}_s \}$ from the face, body, gesture, posture, and scene based on their importance. Specifically,
we design one shared query vector $ \bm{q} \in \mathbb{R}^{d \times 1}$ to obtain the attention values $\psi_{ta} $ as follows:
\begin{equation}
\small
\psi_{ta} = \bm{q}^{T}\cdot tanh(\bm{W}_{ta}\cdot \bm{F}_{ta} + \bm{b}_{ta} ),
\end{equation}
where $\bm{W}_{ta} \in \mathbb{R}^{d \times d}$ and $\bm{b}_{ta} \in \mathbb{R}^{d \times 1}$ are learnable parameters. 
Immediately, the attention values $\psi_{ta} $ are normalized with the softmax function to obtain the final weights:
\begin{equation}
\small
 \gamma _{ta} = \frac{exp(\psi_{ta})}{\sum_{ta\in \{f,b,g,p,s\}}^{} exp(\psi_{ta})}.
\end{equation}
The process provides optimal fusion weights for each feature to highlight the powerful features while suppressing the weaker ones.
The final representation  $\bm{Z}_{fin} \in \mathbb{R}^{d}$ is obtained by the weighted summation:
\begin{equation}
\small
 \bm{Z}_{fin} = \sum_{ta\in \{f,b,g,p,s\}}^{} \gamma_{ta} \odot \bm{F}_{ta}.
\end{equation}

\begin{table}[t]
\setlength{\tabcolsep}{7pt}
\centering
\caption{Configuration for input streams. \textbf{C}: channels; \textbf{F}: frames; \textbf{H}: height; \textbf{W}: width; \textbf{K}: keypoint number; \textbf{P}: human number.}
\resizebox{0.45\textwidth}{!}{%
\begin{tabular}{c|c|c}
\toprule
Stream  & Modality          & Configuration \\ \midrule
Face    & RGB               & 3 (C)$\times$16 (F)$\times$64 (H)$\times$64 (W)    \\
Body    & RGB               & 3 (C)$\times$16 (F)$\times$112 (H)$\times$112 (W)  \\
Gesture & Skeleton Keypoint & 3 (C)$\times$16 (F)$\times$42 (K)$\times$1 (P)     \\
Posture & Skeleton Keypoint & 3 (C)$\times$16 (F)$\times$26 (K)$\times$1 (P)    \\
Scene   & RGB               & 3 (C)$\times$64 (F)$\times$224 (H)$\times$224 (W)  \\ \bottomrule
\end{tabular}
}
\label{config}
\vspace{-0.3cm}
\end{table}
\noindent\textbf{Cross-attention Fusion Module}.
The core idea of this module is to learn pragmatic representations via fine-grained information interaction. We utilize cross-attention to achieve potential adaption from the concatenated source feature $\bm{F}_{so} = [\bm{h}_{f}, \bm{h}_{b}, \bm{h}_{g}, \bm{h}_{p}, \bm{h}_{s}] \in \mathbb{R}^{5d}$ to the target features $\bm{F}_{ta}$ to reinforce each target feature effectively.
Inspired by the self-attention~\cite{vaswani2017attention}, we embed $\bm{F}_{ta}$ into a space denoted as $\mathcal{Q}_{ta}=BN\left (\bm{F}_{ta} \right )\bm{W}_{\mathcal{Q}_{ta}} $, while embedding $\bm{F}_{so}$  into two spaces denoted as $\mathcal{G}_{so}=BN\left (\bm{F}_{so} \right )\bm{W}_{\mathcal{G}_{so}}$ and $\mathcal{U}_{so}=BN\left (\bm{F}_{so} \right )\bm{W}_{\mathcal{U}_{so}} $, respectively. $\bm{W}_{\mathcal{Q}_{ta}}  \in \mathbb{R}^{d \times d}$, $ \{\bm{W}_{\mathcal{G}_{so}}$, $\bm{W}_{\mathcal{U}_{so}} \} \in \mathbb{R}^{5d \times 5d}$ are embedding weights and $BN$ means the batch normalization.
Formally, the cross-attention feature interaction is expressed as follows:
\begin{equation}
\bm{F}_{so\rightarrow ta} = softmax(\mathcal{Q}_{ta} \mathcal{G}_{so}^{T} ) \mathcal{U}_{so} \in \mathbb{R}^{d}.
\end{equation}
Subsequently, the forward computation is expressed as:
\begin{align}
\small
\bm{Z}_{ta} &= BN(\bm{F}_{ta})+\bm{F}_{so\rightarrow ta},\\
\bm{Z}_{ta} &= f_{\delta}(\bm{F}_{ta})+ \bm{Z}_{ta},
\end{align}
where $f_{\delta }(\cdot)$ is the feed-forward layers parametrized by $\delta$, and $\bm{Z}_{ta} \in \{ \bm{Z}_{f}, \bm{Z}_{b}, \bm{Z}_{g}, \bm{Z}_{p}, \bm{Z}_{s} \} \in \mathbb{R}^{d}$.
The reinforced target features $\bm{Z}_{ta}$ are concatenated to get the final representation $\bm{Z}_{fin} \in \mathbb{R}^{d}$ via dense layers.

Finally, four fully connected layers with the task-specific number of neurons are introduced after $\bm{Z}_{fin}$.

\noindent\textbf{Learning Strategies}.
The standard cross-entropy losses are adopted as $ \mathcal{L}_{task}^{k} = - \frac{1}{n} \sum_{i=1}^{n} y_i^k \cdot log \hat{y}_i^k $ for the four classification tasks, where $y_i^{k}$ is the ground truth of the $k$-th task and $n$ is the number of samples in a batch. The total loss is computed as $ \mathcal{L}_{total} = \sum_{k=1}^{4} \lambda_{k} \mathcal{L}_{task}^{k} $, where $\lambda_{k}$ is the trade-off weight. To seek a suitable balance among multiple tasks, we introduce the dynamic weight average~\cite{liu2019end} to adaptively update the weight $\lambda_{k}$ of each task at each epoch.

\section{Experiments}
\subsection{Data Processing}
The input streams are selected from uniform temporal position sampling in synchronized video clips and skeleton sequences, resulting in every 16-frame sample for face, body, gesture, and posture data. To learn the scene semantics efficiently, we merge the sampled clips from the four whole views to produce each 64-frame scene data. Each sample is flipped horizontally and vertically with a 50\% random probability for data augmentation. For the left-right-hand keypoints, we create a link between joints \#94 and \#115 to form an overall gesture topology for processing by a single ST-GCN~\cite{yan2018spatial}. The detailed input configurations for the different streams in each sample are shown in Table~\ref{config}. 

\subsection{Implementation Details}
\noindent\textbf{Experimental Setup}. The whole framework is built on the PyTorch-GPU~\cite{paszke2019pytorch} using four Nvidia Tesla V100 GPUs.
The AdamW~\cite{loshchilov2017decoupled} optimizer is adopted for network optimization with an initial learning rate of 1e-3 and a weight decay of 1e-4. For a fair comparison, the uniform batch size and epoch across models are set to 16 and 30, respectively.
The output dimension $d$ of all models is converted to 128 by minor structural adjustments. In practice, all the hyper-parameters are determined via the validation set. 
Our cross-attention fusion module is the default fusion strategy.

\noindent\textbf{Evaluation Metric}.
We measure recognition performance by classification accuracy (Acc) and weighted F1 score (F1). Considering the demand for practicality~\cite{kotseruba2022attention} in DMS, we provide three-category evaluations of polar emotions and two-category evaluations of abnormal behaviors in the main comparison. Please refer to the \textit{supplementary} for the new taxonomy. The corresponding metrics are the coarse-grained accuracy (CG-Acc) and the F1 score (CG-F1).   

\begin{table}[t]
\setlength{\tabcolsep}{7pt}
\centering
\caption{Experimental results for different streams/modalities. Only weighted F1 scores are reported due to similar results to Acc.}
\resizebox{\linewidth}{!}{%
\begin{tabular}{ccccc|cccc}
\toprule
\multicolumn{5}{c|}{Stream/Modality}    & DER            & DBR            & TCR            & VCR            \\ \midrule
Face & Body & Gesture & Posture & Scene & F1             & F1             & F1             & F1             \\ \midrule
\CheckmarkBold   &     &        &        &      & 66.41          & 51.07          & 48.51          & 41.69          \\
    & \CheckmarkBold   &        &        &      & 63.93          & 62.38          & 55.47          & 50.01          \\
    &     & \CheckmarkBold      &        &      & 52.21          & 57.97          & 50.74          & 58.26          \\
    &     &        & \CheckmarkBold      &      & 65.52          & 63.15          & 55.28          & 47.32          \\
    &     &        &        & \CheckmarkBold    & 49.75          & 45.68          & 86.33          & 75.84          \\ \midrule
\CheckmarkBold   & \CheckmarkBold   &        &        &      & 67.34          & 62.93          & 59.05          & 52.97          \\
\CheckmarkBold   & \CheckmarkBold   & \CheckmarkBold      &        &      & 67.88          & 65.42          & 65.18          & 64.40           \\
\CheckmarkBold   & \CheckmarkBold   & \CheckmarkBold      & \CheckmarkBold      &      & 70.27          & 66.84          & 73.63          & 67.54          \\ 
\CheckmarkBold   & \CheckmarkBold   & \CheckmarkBold      & \CheckmarkBold      & \CheckmarkBold    & \textbf{70.82} & \textbf{67.13} & \textbf{89.98} & \textbf{79.66} \\ \bottomrule
\end{tabular}
}
\label{ab1}
\vspace{-0.4cm}
\end{table}

\subsection{Experimental Results and Analyses}

\noindent\textbf{Main Performance Comparison}.
As shown in Table~\ref{tab2}, we comprehensively report the comparison results of different baseline models combined in the three learning patterns.
The following are some key observations.
(\textbf{\rmnum{1}}) The overall performance (Acc/F1) of the DER, DBR, TCR, and VCR tasks approaches only around 72\%, 67\%, 89\%, and 79\%, respectively, which still leaves considerable improvement room.
(\textbf{\rmnum{2}}) The results in 3D and 2D + Timing patterns are generally better than those in 2D for all four tasks, demonstrating that considering temporal information can help improve perception performance. This makes sense as sequential modeling captures the rich dynamical clues among frames. For instance, the TransE-based Experiment (9) shows a significant gain of 3.50\% and 6.15\% in Acc and F1 on the DBR task compared to its 2D version (4).
(\textbf{\rmnum{3}}) In the 3D pattern, resource-efficient model combinations can also achieve competitive or even better results compared to dense structures, as in Experiments (11, 13). This finding inspires researchers to consider the performance-efficiency trade-off when selecting suitable DMS models.
(\textbf{\rmnum{4}}) Experiments (1, 6) reveal that the rich scene semantics in the Places365 dataset~\cite{zhou2017places} facilitates capturing valuable context prototypes from the pre-trained backbone, leading to better performance on the TCR and VCR tasks.

\noindent\textbf{Importance of Distinct Streams/Modalities}.
To investigate the impact of distinct streams/modalities, we conduct experiments using the performance-balanced combination (13) with increasing inputs. Table~\ref{ab1} shows the following interesting findings. 
(\textbf{\rmnum{1}}) For isolated inputs, the scene stream provides the most beneficial visual clues for determining traffic context and vehicle condition. The body and posture modalities are more competitive on the DER and DBR tasks, indicating that bodily expressions can convey critical intent information. The observation is consistent with psychological research~\cite{cornelius1996science,yang2022emotion}.
(\textbf{\rmnum{2}}) With the progressive increase in information channels, various driver-based characteristics contribute to emotion and behavior understanding.
(\textbf{\rmnum{3}}) The body and posture streams bring meaningful gains of 10.54\% and 8.45\% to the TCR task compared to the preceding one, showing that driver attributes are potentially related to the traffic context. For example, drivers usually change their gait during \textit{traffic jam} to perform irrelevant operations~\cite{li2021survey}.
(\textbf{\rmnum{4}}) The gesture modality promisingly improves the VCR task's result by 11.43\% compared to the preceding one. A reasonable interpretation is that vehicle states highly correlate with specific hand motions, \eg, the two hands generally cross when the vehicle is \textit{turning}.

\begin{table}[t]
\caption{Experimental results for different perception tasks. ``2DT'' means ``2D + Timing'' pattern. ``w/o'' stands for the without.}
\resizebox{\linewidth}{!}{%
\begin{tabular}{cl|c|cc|cc|cc|cc}
\toprule
\multicolumn{2}{c|}{\multirow{2}{*}{Config}}     & \multirow{2}{*}{Pattern} & \multicolumn{2}{c|}{DER} & \multicolumn{2}{c|}{DBR} & \multicolumn{2}{c|}{TCR} & \multicolumn{2}{c}{VCR} \\ \cline{4-11} 
\multicolumn{2}{c|}{}                            &                          & Acc         & F1         & Acc         & F1         & Acc         & F1         & Acc        & F1         \\ \midrule
\multicolumn{2}{c|}{\multirow{3}{*}{Full Tasks}} & 2D                       & \textbf{71.26}       & \textbf{68.71}      & \textbf{65.35}       & \textbf{63.29}      & \textbf{83.74}       & \textbf{81.28}      & \textbf{77.12}     & \textbf{75.23}      \\
\multicolumn{2}{c|}{}                            & 2DT                      & \textbf{70.83}       & \textbf{67.14}     & \textbf{67.32}       & \textbf{64.45}      & \textbf{90.54}       & \textbf{89.66}      & \textbf{79.97}      & \textbf{77.94}      \\
\multicolumn{2}{c|}{}                            & 3D                       & \textbf{74.87}       & \textbf{72.56}      & \textbf{65.18}       & \textbf{63.24}      & \textbf{92.12}       & \textbf{91.81}      & \textbf{78.81}      & \textbf{76.91}      \\ \midrule
\multicolumn{2}{c|}{\multirow{3}{*}{w/o DER}}    & 2D                       & -           & -          & 63.13       & 60.96      & \textbf{84.55}       & \textbf{81.79}      & 77.07       & 75.16      \\
\multicolumn{2}{c|}{}                            & 2DT                      & -           & -          & 65.08       & 62.72      & 90.20        & 89.27      & 79.86      & 77.85      \\
\multicolumn{2}{c|}{}                            & 3D                       & -           & -          & 63.47       & 61.35      & 91.86       & 90.74      & \textbf{78.85}      & \textbf{76.94}      \\ \midrule
\multicolumn{2}{c|}{\multirow{3}{*}{w/o DBR}}    & 2D                       & 70.29       & 67.44      & -           & -          & 80.92       & 78.66      & 74.58      & 72.92      \\
\multicolumn{2}{c|}{}                            & 2DT                      & 68.03       & 64.58      & -           & -          & 87.22       & 86.51      & 77.51      & 75.67      \\
\multicolumn{2}{c|}{}                            & 3D                       & 72.54        & 69.62      & -           & -          & 89.61       & 89.37      & 76.42      & 74.55      \\ \midrule
\multicolumn{2}{c|}{\multirow{3}{*}{w/o TCR}}    & 2D                       & 71.23       & 68.67      & 64.42       & 62.36      & -           & -          & 76.72      & 74.60       \\
\multicolumn{2}{c|}{}                            & 2DT                      & \textbf{70.95}       & \textbf{67.22}     & 65.18       & 62.33      & -           & -          & 77.54      & 75.46      \\
\multicolumn{2}{c|}{}                            & 3D                       & 74.61       & 72.28      & 65.15       & 63.19      & -           & -          & 78.02      & 76.15      \\ \midrule
\multicolumn{2}{c|}{\multirow{3}{*}{w/o VCR}}    & 2D                       & \textbf{71.43}       & \textbf{69.17}      & 63.24       & 63.15      & 83.65        & 81.14      & -          & -          \\
\multicolumn{2}{c|}{}                            & 2DT                      & 70.79       & 67.02      & 66.11       & 63.04      & \textbf{91.23}       & \textbf{90.28}      & -          & -          \\
\multicolumn{2}{c|}{}                            & 3D                       & 74.57       & 72.18      & 64.76       & 62.75      & 92.04       & 91.75      & -          & -          \\ \bottomrule
\end{tabular}
}
\label{ab2}
\vspace{-0.4cm}
\end{table}

 \begin{figure*}[t]
\centering
\includegraphics[width=\linewidth]{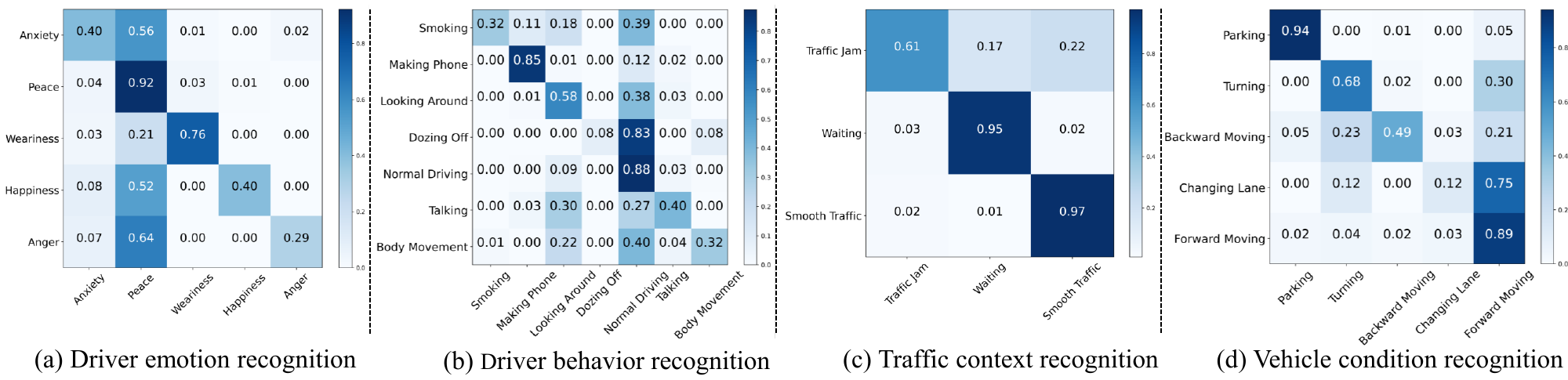}
\caption{Confusion matrices for the best model performance from the four tasks.
  }
\label{matrix}
\vspace{-0.4cm}
\end{figure*}

\noindent\textbf{Necessity of Different Perception Tasks}.
In Table~\ref{ab2}, we select the Experiments (2, 6, 20) to verify the necessity of different perception tasks in the three patterns. Each task is removed separately to observe the performance variation of the other tasks. We have the following insights.
(\textbf{\rmnum{1}}) When all four tasks are present simultaneously, the best overall results are achieved across different patterns, confirming that these tasks can synergistically achieve holistic perception.
(\textbf{\rmnum{2}}) The interaction between the DER and DBR tasks is more significant, implying a solid mapping between driver-based representations.
For instance, negative emotional states (\eg, \textit{anxiety}) are more likely to induce secondary behaviors (\eg, \textit{looking around}) and cause accidents~\cite{jeon2014effects}.
(\textbf{\rmnum{3}}) The DBR task offers valuable average gains of 2.88\%/2.74\% and 2.46\%/2.31\% for the TCR and VCR tasks regarding Acc/F1, respectively, indicating a beneficial correlation between the driver's state inside the vehicle and the traffic scene outside.

\begin{table}[t]
\setlength{\tabcolsep}{7pt}
\centering
\caption{Experimental results for multiple views and different fusion strategies. ``w/o'' stands for the without. }
\resizebox{\linewidth}{!}{%
\begin{tabular}{clcccc}
\toprule
\multicolumn{2}{c|}{\multirow{2}{*}{Config}}  & DER   & DBR   & TCR   & VCR   \\ \cline{3-6} 
\multicolumn{2}{c|}{}                         & Acc   & Acc   & Acc   & Acc   \\ \midrule
\multicolumn{2}{c|}{Full Framework}                     & \textbf{70.11} & \textbf{66.52} & \textbf{88.51} & \textbf{81.12} \\ \midrule
\multicolumn{6}{c}{Effectiveness of Multiple Views}                          \\ \midrule
\multicolumn{2}{c|}{w/o Inside View}          & 68.08 & 64.41 & \textbf{88.54} & 80.64 \\
\multicolumn{2}{c|}{w/o Front View}           & 69.85 & 65.67 & 76.80  & 76.72 \\
\multicolumn{2}{c|}{w/o Left View}            & \textbf{70.11} & 66.48 & 84.39 & 71.43 \\
\multicolumn{2}{c|}{w/o Right View}           & 70.06 & \textbf{66.55} & 85.26 & 72.55 \\ \midrule
\multicolumn{6}{c}{Impact of Different Fusion Strategies}                             \\ \midrule
\multicolumn{2}{c|}{Adaptive Fusion Module (ours)} & \textbf{70.20}  & 65.36 & \textbf{88.57} & 80.34 \\
\multicolumn{2}{c|}{Feature Summation}                & 66.85 & 64.53 & 85.19 & 77.56 \\
\multicolumn{2}{c|}{Feature Concatenation}            & 68.33 & 64.79 & 87.05 & 78.02 \\
\bottomrule
\end{tabular}
}
\label{ab3}
\vspace{-0.4cm}
\end{table}

\noindent\textbf{Effectiveness of Multiple Views}.
From Table~\ref{ab3} (\textit{top}), we employ the Experiment (15) to evaluate the effectiveness of multiple views.
(\textbf{\rmnum{1}}) We find that the DER and DBR tasks benefit mainly from the inside view, as the interior scene provides necessary recognition clues, such as driver-related information and vehicle internals. 
The inside view brings gains (Acc) of 2.03\% and 2.11\% for driver emotion and behavior understanding, respectively.
(\textbf{\rmnum{2}}) The three out-of-vehicle views provide indispensable contributions to the TCR and VCR tasks, as they contain perceptually critical traffic context semantics.
(\textbf{\rmnum{3}}) The multi-view setting of AIDE achieves an overall better performance across tasks via complementary information sources.

\noindent\textbf{Impact of Fusion Strategies}.
We explore the impact of different fusion strategies in Table~\ref{ab3} (\textit{bottom}).
(\textbf{\rmnum{1}}) Our adaptive fusion achieves a noteworthy performance compared to the default cross-attention fusion, indicating that both fusion paradigms are superior and usable.
(\textbf{\rmnum{2}}) Feature summation and concatenation may introduce redundant information leading to poor results and sub-optimal solutions.

\noindent\textbf{Analysis of Confusion Matrices}.
For the different classification perception tasks, Figure~\ref{matrix} shows the confusion matrices under the best results in each task to analyze the performance of each class. (\textbf{\rmnum{1}}) Due to the interference of the long-tail distribution (Figure~\ref{count}), some head classes are usually confused with other classes, such as ``\textit{peace}'' from the DER task in Figure~\ref{matrix}(a) and ``\textit{forward moving}''  from the VCR task in Figure~\ref{matrix}(d).
Moreover, the sparse tail samples lead to inadequate learning of class-specific representations, such as ``\textit{dozing off}''  from the DBR task in Figure~\ref{matrix}(b). These phenomena are inevitable because the driver remains safely driving for long periods of time in most naturalistic scenarios.
(\textbf{\rmnum{2}}) In Figure~\ref{matrix}(c), ``\textit{traffic jam}'' creates evident confusion with the other classes. The possible reason is that the rich information from distinct out-of-vehicle views unintentionally exaggerates the scene context clues.

\section{Conclusion and Discussion}
In this paper, we present the AssIstive Driving pErception Dataset (AIDE) to facilitate the development of next-generation Driver Monitoring Systems (DMS) in a perceptually comprehensive manner. With its multi-view, multi-modal, and multi-tasking advantages, AIDE achieves effective collaborative perception among driver emotion, behavior, traffic context, and vehicle condition. In this case, we evaluate extensive model combinations and component ablations in three pattern frameworks to systematically demonstrate the importance of AIDE. 

AIDE potentially provides a valuable resource for studying distinct driving recognition tasks with imbalanced data.
Furthermore, we empirically suggest that future research could be considered as follows:
(\textbf{\rmnum{1}}) Mining causal effects among driving dynamics inside and outside the vehicle to disentangle data distribution gaps in different tasks.
(\textbf{\rmnum{2}}) Developing unified resource-efficient structures to achieve performance-efficiency trade-offs in the pragmatic DMS. 

\section*{Acknowledgements}
We thank the anonymous reviewers for providing constructive discussions and suggestions.
This work is supported in part by the National Key R\&D Program of China (2021ZD0113503), in part by the Shanghai Municipal Science and Technology Major Project (2021SHZDZX0103), and in part by the China Postdoctoral Science Foundation under Grant (BX20220071, 2022M720769).

{\small
\bibliographystyle{ieee_fullname}
\bibliography{iccv}
}

\end{document}